\documentclass[letterpaper]{article}
\usepackage{palatepreprint}
\usepackage[hyphens]{url}
\usepackage{graphicx}
\usepackage{natbib}
\usepackage{caption}
\usepackage{booktabs}
\title{PALATE: Personalized Aesthetic Learning through Adaptive Taste Evolution \\for Multi-User Portrait Retouching}

\author{
    Jingxuan Wang\textsuperscript{\rm 1,\rm 4},
    Yifan Mei\textsuperscript{\rm 2,\rm 4},
    Yuxia Niu\textsuperscript{\rm 3,\rm 4},
    Chaowan Jiao\textsuperscript{\rm 2,\rm 4},
    Qijin Shen\textsuperscript{\rm 4}\corresponding
}

\affiliations{
    \textsuperscript{\rm 1}Harbin Institute of Technology\quad
    \textsuperscript{\rm 2}Xiamen University\quad
    \textsuperscript{\rm 3}Central South University\quad
    \textsuperscript{\rm 4}MT Lab, Meitu Inc.
}

\begin{document}
\maketitle
\begin{abstract}
Automatic portrait retouching has advanced rapidly, yet its objective is inherently subjective: the same portrait admits multiple professionally valid results, and users disagree about which one is best. Most existing methods optimize a population-level aesthetic standard and therefore cannot capture individual taste, while fine-tuning a separate editing model for every user incurs prohibitive training, storage, and data costs. We propose PALATE, a shared reward-evolution framework that keeps the image editor fixed and instead personalizes the selection among retouched candidates of the same source portrait. PALATE decomposes the reward for each user into a global backbone shared by all users, category-level residuals shared by aesthetically similar users, and a lightweight user adapter, with anti-collapse regularizers keeping the three levels complementary.A cyclic dual-level distillation scheme first distills user-specific preferences into category rewards and then consolidates the resulting category-level knowledge into the global backbone, which is redistributed to initialize the next evolution round. In this way, the shared initialization improves progressively across rounds, enabling unseen users to be calibrated from only a few rankings. On expert-retouched candidates from PPR10K with held-out users and held-out images, PALATE attains 72.83\% pairwise preference-prediction accuracy, surpassing all reward, aesthetic, and image-quality baselines, of which the strongest, PickScore, reaches 58.06\%. Each new user costs only 512 bytes of user-specific parameters and millisecond-level scoring.
\end{abstract}

\section{Introduction}

A single portrait admits multiple professionally valid retouchings---with warmer or cooler skin tones, stronger or lighter smoothing, and restrained or expressive color---but the preferred result varies across users. This subjectivity becomes increasingly important as portrait images serve as a primary means of self-presentation on visual social media, where users expect retouching results to reflect their individual aesthetics rather than a universal standard.
Existing solutions force a trade-off between usability and personalization: professional tools such as Photoshop offer fine-grained control but require substantial expertise, whereas consumer presets are easy to use but produce homogeneous results that fail to accommodate individual preferences. A seemingly direct remedy is to adapt the underlying editing model to each user \citep{zhu2025pif} with parameter-efficient modules such as adapters \citep{houlsby2019adapter} or LoRA \citep{hu2022lora}. For a large and growing user population, however, separately training, storing, and deploying user-specific editors is prohibitively expensive, and reliable adaptation demands considerable per-user data and computation.

We instead shift personalization from the editing model to the reward model. Rather than fine-tuning an editor for every user, we learn and iteratively refine a shared reward initialization that accumulates preference knowledge from historical users and enables rapid adaptation to unseen ones. Achieving this is nontrivial: user preferences mix transferable regularities with irreducible individual differences, so a purely global reward washes out personal taste, whereas independently trained user rewards cannot reuse knowledge across users.

\begin{figure*}[t]
\centering
\includegraphics[width=0.97\textwidth]{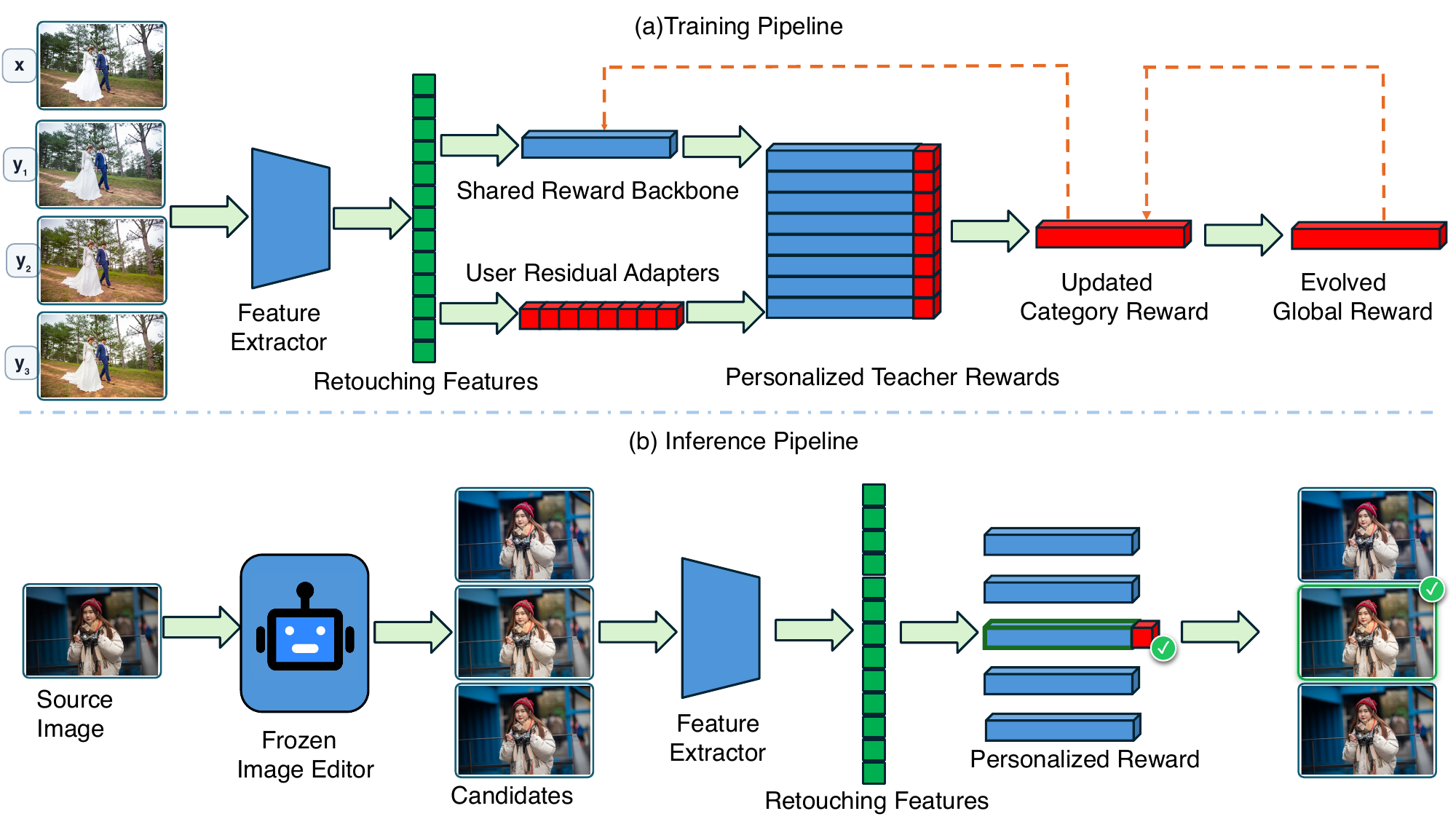}
\caption{Overview of PALATE.
(a) During training, a source portrait and its retouched candidates
are encoded into retouching features. A shared reward backbone and
user residual adapters jointly form personalized teacher rewards.
User-level preference knowledge is first distilled within each
aesthetic category to update the corresponding category reward,
after which category rewards are consolidated into an evolved global
reward. The evolved rewards are redistributed to initialize the next
round of category modeling and user adaptation.
(b) During inference, a frozen image editor generates multiple
retouched candidates for a source portrait, and a few-shot calibrated
personalized reward ranks the candidates and selects the result that
best matches the target user's preference.}
\label{fig:framework}
\end{figure*}

We therefore propose PALATE, a hierarchical reward-learning framework with cyclic dual-level distillation for personalized portrait-retouching candidate selection, illustrated in Figure~\ref{fig:framework}. PALATE decomposes the personalized reward into three levels: a global backbone that captures aesthetic knowledge shared across users, category-level residuals that model preference patterns shared by aesthetically similar users, and lightweight user adapters that encode individual variations. To keep this hierarchy meaningful, a dual-constraint mechanism keeps each adapter informative and distinguishable in the shared preference space, preventing user representations from collapsing toward a few dominant solutions.

As shown in the upper part of Figure~\ref{fig:framework}, preference knowledge is consolidated through two distillation stages. User-level rewards are first distilled into their category models, turning individual feedback into reusable category-level preference knowledge; category knowledge is then consolidated into the global backbone, which is redistributed to initialize the next round of category and user adaptation. Through this cycle, the shared reward continually accumulates and reuses multi-user knowledge instead of repeatedly adapting from a static initialization.

For an unseen user, the evolved backbone provides a robust initialization: the best-matching category reward is selected from a small support set, and only a lightweight user adapter is optimized. PALATE thus serves every user with one frozen editor, one cyclically evolving shared reward, and a few hundred bytes of per-user state.

Our main contributions are as follows:

\begin{enumerate}
\item We propose a global--category--user hierarchical reward architecture that separates shared aesthetic knowledge, category-level preference patterns, and individual residuals.
\item We introduce a dual-constraint mechanism that keeps user adapters informative yet mutually distinguishable, mitigating representation collapse across users.
\item We develop a cyclic dual-level distillation scheme that consolidates knowledge from users to categories and from categories to the global backbone, yielding progressively stronger shared initializations for few-shot adaptation to unseen users.
\end{enumerate}
\section{Related Work}

\subsection{Portrait Retouching and Controllable Image Editing}

Learning-based photo retouching was initially formulated as learning a
mapping from source images to expert references, as exemplified by
FiveK \citep{bychkovsky2011fivek}. PPR10K
\citep{liang2021ppr10k} extends this paradigm to portraits with
human-region masks and multiple professional retouches. Subsequent
methods improve efficiency, interpretability, and regional control
through bilateral transforms, white-box operators, local parametric
filters, learnable lookup tables, and region-specific color filters
\citep{gharbi2017hdrnet,hu2018exposure,moran2020deeplpf,
zeng2022lut,he2020csrnet,ouyang2023rsfnet,xie2023bpfre}.

More recent studies shift attention from fixed enhancement mappings to
flexible control. Conditional Image Repainting methods progressively
explore semantic bridging, unified multi-condition fusion,
illumination-aware repainting, and open condition mixtures
\citep{weng2020cirbridge,sun2022unicorn,tang2023luminaire,
weng2024cir,weng2025opencir}. Instruction-guided editors such as
InstructPix2Pix \citep{brooks2023instructpix2pix} and MagicBrush
\citep{zhang2023magicbrush} translate natural-language requests
into visual modifications, while HIVE \citep{zhang2024hive} further
incorporates human feedback to align editing results. Affective Image
Filter \citep{weng2023aif} and its generative-prior extension
\citep{zhang2026aifd} map emotional text to image adjustments, while
AIEdiT \citep{zhang2026aiedit} controls multiple affective factors
from fine-grained descriptions. PIF \citep{zhu2025pif} instead
extracts and transfers photographic styles from reference images.
Despite their flexibility, these methods rely on explicit conditions
to manipulate images, rather than learning a persistent user-specific
reward for selecting among professionally retouched candidates.

\subsection{Personalized Visual Preference}

Existing preference models differ in whose judgment they represent.
ImageReward \citep{xu2023imagereward} and HPS
\citep{wu2023hps} aggregate comparisons into general
human-preference scores, while Pick-a-Pic
\citep{kirstain2023pickapic} and HPS v2
\citep{wu2023hpsv2} enlarge the crowd-sourced preference data and
evaluation protocols available for training such models. MPS
\citep{zhang2024mps} further separates preference into multiple
evaluation dimensions. None of these general-purpose models maintains
a persistent representation of individual taste.

Personalized aesthetic models address this limitation through residual
corrections \citep{ren2017personalized}, rich user and image
attributes \citep{yang2022para}, or user-adaptive meta-learning
\citep{zhu2022blgpiaa}. Recent preference-learning methods also adapt
generative models from limited pairwise feedback
\citep{dang2025personalized}, while variational preference learning
\citep{poddar2024vpl} and low-rank reward modeling
\citep{bose2025lore} represent heterogeneous users with latent
variables or user-specific combinations of shared reward components.
Most nevertheless connect a shared model directly to individual
parameters, without explicitly capturing preference regularities
recurring across aesthetically similar users.

\subsection{Multi-User Knowledge Transfer}

Knowledge distillation \citep{hinton2015distilling} transfers
predictive structure between models, including intermediate
representations rather than only final predictions, as in FitNets
\citep{romero2015fitnets}. Deep Mutual Learning
\citep{zhang2018dml} enables collaborative supervision, whereas
Born-Again Networks \citep{furlanello2018born} and Knowledge
Evolution \citep{taha2021knowledge} repeatedly refine knowledge
across successive generations. Multi-teacher methods further combine
intermediate teacher assistants \citep{son2021densely} or dynamically
selected teacher groups \citep{choi2023orc} to improve knowledge
transfer.

Beyond centralized distillation, FedDF \citep{lin2020feddf}
consolidates heterogeneous client predictions into a global model,
and multi-teacher distillation \citep{wen2024cilmtd} has been used to
preserve knowledge across incremental learning stages. These methods
generally seek a stronger consensus student. In personalized
aesthetics, however, disagreement may encode meaningful taste rather
than noise: direct aggregation can erase user differences, whereas
isolated user models cannot convert recurring preferences into shared
knowledge. PALATE addresses this combined gap through hierarchical
cyclic reward transfer that preserves group-level variation while
improving the shared model across rounds.

\section{Method}
PALATE is a personalized candidate-selection framework for portrait retouching. As summarized in Figure~\ref{fig:framework}, it models retouching preferences at three granularities: a global backbone capturing broadly transferable retouching knowledge, a category residual preserving the distinctive patterns of each preference category, and a lightweight user adapter calibrating the reward to an individual user. Anti-collapse regularizers at the category and user levels keep the learned representations from degenerating into similar solutions, so each component retains complementary preference knowledge. A cyclic dual-level distillation strategy then consolidates preference knowledge across the user, category, and global levels, letting the reward hierarchy evolve over successive rounds.

\subsection{Ranking-Based Preference Supervision}
For each source portrait \(x\), PPR10K \citep{liang2021ppr10k}
provides three expert-retouched candidates
\(\mathcal{Y}_x=\{y_1,y_2,y_3\}\).
A user \(u\) from preference category \(c(u)\) ranks them as
\(y_{\pi_1}\succ_{u,x}y_{\pi_2}\succ_{u,x}y_{\pi_3}\).
Each ranking induces three pairwise preference tuples
\(z=(x,y^{+},y^{-},u,c(u))\) with \(y^{+}\succ_{u,x}y^{-}\),
and the resulting collection \(\mathcal{D}_{\mathrm{pref}}\)
constitutes our supervision. Ranking triplets are a
low-burden interface: ordering three candidates takes
seconds, yet every ranking yields three
mutually consistent pairwise constraints. All comparisons induced by
the same \((u,x)\) ranking are kept in the same split to
avoid information leakage.

\subsection{Hierarchical Residual Reward Modeling}
Given a source portrait \(x\) and a retouched candidate \(y\),
the global backbone \(F_{\theta_G^t}\) maps the feature
vector \(\Phi(x,y)\) to a scalar reward \(G^t(x,y)\) and an
\(\ell_2\)-normalized preference representation
\(h^t(x,y)\), where \(t\) denotes the evolution round and
\(\Phi\) extracts portrait-appearance and source--edit
features. The personalized
reward for user \(u\) is
\begin{equation}
R_u^t(x,y)
=
G^t(x,y)
+
C_{c(u)}^t(x,y)
+
e_u^{\top}h^t(x,y),
\label{eq:hierarchical_reward}
\end{equation}
where \(C_{c(u)}^t\) is the residual reward of category \(c(u)\)
and \(e_u\) is a low-dimensional user residual; at deployment,
PALATE returns \(\mathop{\rm arg\,max}_{y\in\mathcal{Y}_x}R_u^t(x,y)\).
For a preference tuple
\(z_n=(x_n,y_n^+,y_n^-,u_n,c_n)\), let
\(\delta_f(z_n)\) denote the pairwise margin predicted by reward
function \(f\). We define the Bradley--Terry ranking loss
\citep{bradley1952rank} and the margin-distillation distance as
\begin{equation}
\begin{array}{rcl}
\delta_f(z_n)
&=&
f(x_n,y_n^+)-f(x_n,y_n^-),
\\\\
\mathcal{L}_{\mathrm{rank}}(\mathcal{D};f)
&=&
-\displaystyle\frac{1}{|\mathcal{D}|}
\displaystyle\sum_{z_n\in\mathcal{D}}
\displaystyle\log\sigma\bigl(\delta_f(z_n)\bigr),
\\
\mathrm{D}_{\Delta}(f,g;\mathcal{D})
&=&
\displaystyle\frac{1}{|\mathcal{D}|}
\displaystyle\sum_{z_n\in\mathcal{D}}
\displaystyle\bigl(
\delta_f(z_n)-\delta_g(z_n)
\bigr)^2.
\end{array}
\label{eq:pairwise_objectives}
\end{equation}

\subsection{Hierarchical Residual Geometry Regularization}
The additive reward decomposition alone does not guarantee
a meaningful hierarchy. Shared preference patterns may be
absorbed by lower-level residuals, while user representations
may collapse toward a few dominant directions. We therefore
regularize the residual geometry at both the user and category
levels.

\paragraph{Centered user residuals.}
For each training user \(u\), we maintain \(P\) bootstrap
latent codes \(q_{u,p}\in\mathbf{R}^{d_q}\). Let
\(\mathcal{U}_c\) denote the users assigned to category
\(c\), \(\bar q_c\) their mean latent code, and \(A_c\)
the map from latent codes to the preference space. The user
residual is
\begin{equation}
e_u
=
\frac{s}{P}
\sum_{p=1}^{P}
A_c^{\top}\bigl(q_{u,p}-\bar q_c\bigr),
\label{eq:centered_user}
\end{equation}
where \(s\) controls the residual scale. Category-wise
centering prevents components shared by all users of a
category from being re-encoded by their adapters; it also
gives the hierarchy a clean interpretation: the category
reward captures the collective taste of a group, and each
adapter encodes only a user's deviation from it.
To preserve diversity among user representations, we apply
SIGReg \citep{balestriero2025lejepa} to random one-dimensional
projections. Let \(Z_{c,\ell}\) collect the projections
\(\theta_{\ell}^{\top}(q_{u,p}-\bar q_c)\) of all centered
codes in category \(c\) onto a random unit direction
\(\theta_{\ell}\). The user-level regularizer is
\begin{equation}
\mathcal{L}_{\rm usr}
=
\frac{1}{CL}
\sum_{c=1}^{C}
\sum_{\ell=1}^{L}
D_{\rm EP}
\bigl(
Z_{c,\ell},\mathcal{N}(0,1)
\bigr),
\label{eq:user_sigreg}
\end{equation}
where \(D_{\rm EP}\) denotes the Epps--Pulley discrepancy
\citep{epps1983test}. During
user fitting, the global and category rewards remain fixed,
and the user codes are optimized with
\(\mathcal{L}_{\rm rank}+\lambda_{\rm usr}\mathcal{L}_{\rm usr}\).

\paragraph{Category residual geometry.}
We characterize each category on a fixed training-only anchor
set \(\mathcal{B}=\{(x_m,y_m^{+},y_m^{-})\}_{m=1}^{M}\)
through its residual-margin signature \(v_c\in\mathbf{R}^{M}\),
with entries \([v_c]_m=C_c(x_m,y_m^{+})-C_c(x_m,y_m^{-})\).
Let \(\widetilde v_c\) denote the signature centered across
categories and \(\widehat v_c\) its \(\ell_2\)-normalized
counterpart, stacked as the rows of \(\widehat V\). The
category regularizer is
\begin{equation}
\mathcal{L}_{\rm cat}
=
\frac{1}{C}\sum_{c=1}^{C}
\left[
\tau-\frac{\|\widetilde v_c\|_2}{\sqrt{M}}
\right]_{+}
+
\left\|
\widehat V\widehat V^{\top}-S_C
\right\|_{F}^{2},
\label{eq:category_geometry}
\end{equation}
where \(S_C\) is the Gram matrix of a regular simplex, with diagonal entries equal to \(1\) and
off-diagonal entries equal to \(-1/(C-1)\). The first term
prevents category residuals from vanishing, whereas the
second encourages different categories to occupy distinct
directions. Geometrically, the regular simplex is the
configuration in which \(C\) unit vectors are maximally and
equally separated---the geometry that also emerges in neural
collapse \citep{papyan2020collapse}---so the regularizer
drives the categories toward equally distinct preference
directions without privileging any single group. This regularizer is added to
category-level training with coefficient \(\lambda_a\).
\begin{figure}[t]
\centering
\makebox[\columnwidth][r]{%
    \includegraphics[
        width=0.98\columnwidth
    ]{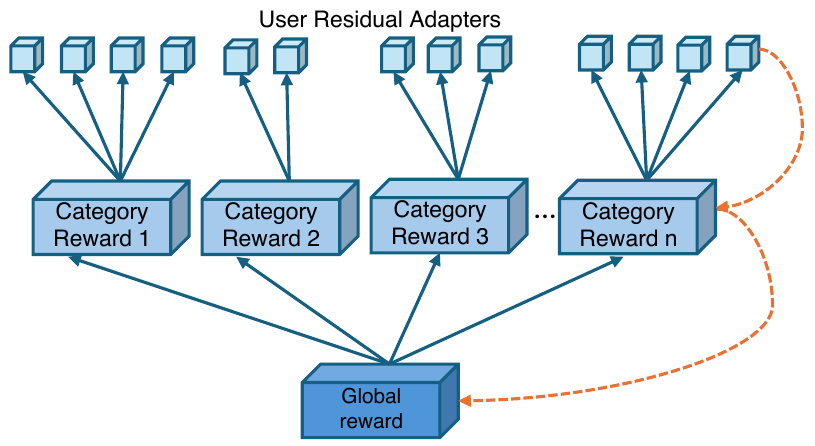}%
}
\caption{
Global--category--user cyclic reward evolution in PALATE.
Shared knowledge is distributed from the global reward to category
rewards and user adapters, while personalized knowledge is distilled
in the reverse direction, from users to categories and then to the
global reward. The evolved global reward is redistributed to
initialize the next evolution round.
}
\label{fig:cyclic_distillation}
\end{figure}
\subsection{Cyclic Dual-Level Reward Distillation}
PALATE evolves the hierarchy cyclically from users to
categories, from categories to the global model, and back to
categories. All three steps instantiate one
\emph{distill-and-anchor} objective. For a student reward
\(f\), a teacher reward \(T\), a sample set
\(\mathcal{D}\), and the student's previous state
\(f^{-}\), we minimize
\begin{equation}
\mathcal{J}
=
\mathcal{L}_{\mathrm{rank}}(\mathcal{D};f)
+
\lambda\,
\mathrm{D}_{\Delta}(f,T;\mathcal{D})
+
\mu\,
\Omega(f,f^{-}),
\label{eq:distill_template}
\end{equation}
where \(\mathcal{L}_{\mathrm{rank}}\) and
\(\mathrm{D}_{\Delta}\) are defined in
Eq.~(\ref{eq:pairwise_objectives}) and \(\Omega\) penalizes
parameter drift from the previous state. The ranking term
anchors learning to observed preferences, margin distillation
transfers the teacher's preference structure, and the drift
penalty stabilizes the cycle; the three transfer steps differ
only in their teacher, sample set, and anchor.

\paragraph{User-to-category distillation.}
With \(G^t\) and \(C_c^t\) fixed, the fitted user residuals
define personalized teachers
\(
T_u^t(x,y)
=
G^t(x,y)
+
C_{c(u)}^t(x,y)
+
e_u^{\top}h^t(x,y).
\)
Each category residual is updated by
\begin{equation}
C_c^{t,+}
=
\mathop{\rm arg\,min}\limits_{C_c}
\frac{1}{|\mathcal{U}_c|}
\sum_{u\in\mathcal{U}_c}
\mathcal{J}\bigl(G^t{+}C_c;T_u^t,\mathcal{D}_{u,c}^{UC}\bigr)
+
\lambda_a\mathcal{L}_{\mathrm{cat}},
\label{eq:user_to_category}
\end{equation}
where the ranking term inside \(\mathcal{J}\) is shared over
the category-level union \(\mathcal{D}_c^{UC}\) and
\((\lambda,\mu)\) instantiate as \((\lambda_u,\lambda_p)\).
Margin distillation thus transfers preference patterns
shared by the users of a category.

\paragraph{Category-to-global consolidation and feedback.}
The evolved backbone
\(G^{t+1}\) minimizes
Eq.~(\ref{eq:distill_template}) with student \(G\), the
\(C\) category teachers \(G^t+C_c^{t,+}\), each weighted by
\(\lambda_c/C\) on \(\mathcal{D}_c^{CG}\), and
drift penalty \(\lambda_g\Omega(G,G^t)\); equal category
weighting prevents categories with more users or samples from
dominating the shared backbone. The category residuals are then
rebased on the evolved global reward: each \(C_c^{t+1}\)
minimizes Eq.~(\ref{eq:distill_template}) with student
\(G^{t+1}+C_c\), teacher \(G^t+C_c^{t,+}\), samples
\(\mathcal{D}_c^{GC}\), weights
\((\lambda,\mu)=(\lambda_b,\lambda_p)\), and
\(\lambda_a\mathcal{L}_{\mathrm{cat}}\) added, which
preserves category-specific margins after the global reference
changes. We carry \(G^{t+1}\) and
\(\mathbf{C}^{t+1}=\{C_c^{t+1}\}_{c=1}^{C}\) into the next
evolution round.
To reduce teacher--student leakage, the source portraits used
for user fitting, user-to-category distillation, and
category-to-global consolidation are assigned to disjoint
training subsets.

\begin{figure*}[t]
\centering
\includegraphics[
    width=0.78\textwidth
]{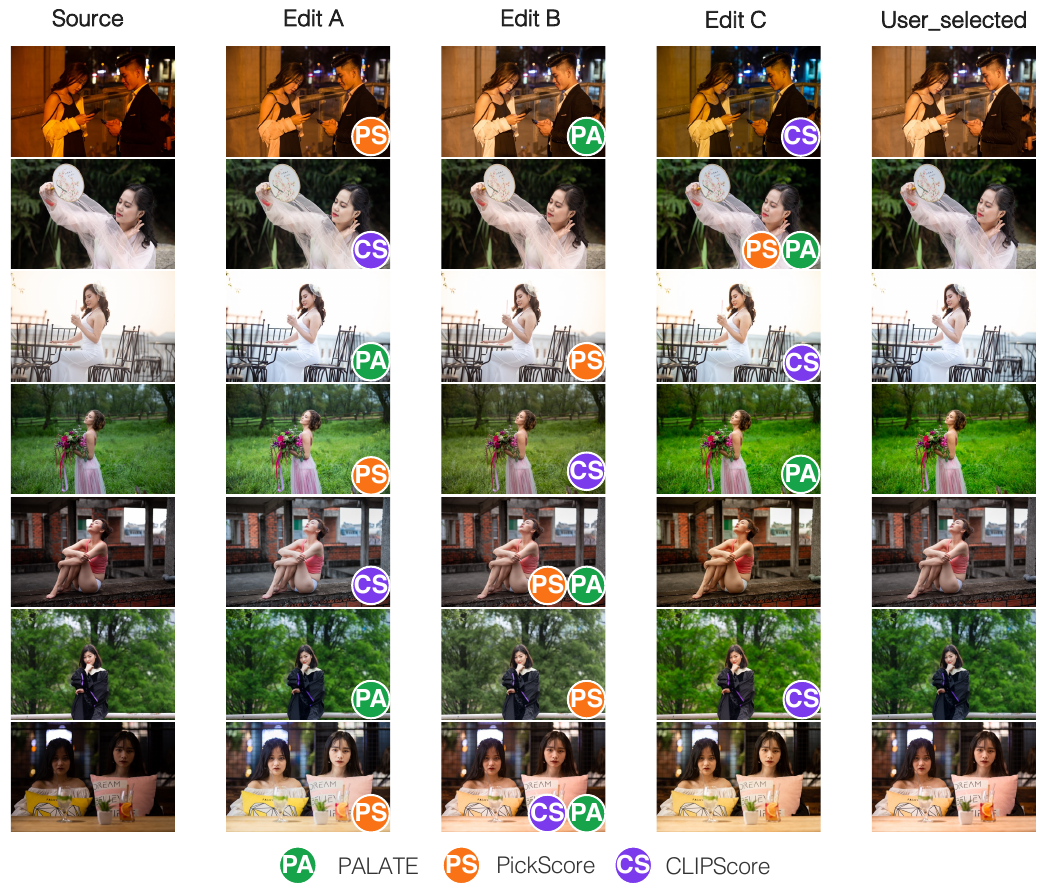}
\caption{
Qualitative comparison of candidate-selection results.
For each source portrait, three expert-retouched candidates are
evaluated by PALATE, PickScore, and CLIPScore.
The rightmost column shows the candidate selected by the user.
}
\label{fig:qualitative_comparison}
\end{figure*}

\subsection{Safe Calibration for New Users}
At test time, the evolved global and category rewards are frozen.
Given a support set \(\mathcal{S}_u\) containing a few
rankings from a new user, PALATE estimates only a
low-dimensional user residual by minimizing the Bradley--Terry
objective in Eq.~(\ref{eq:pairwise_objectives}) with an
\(\ell_2\) penalty:
\begin{equation}
\widehat{e}_u
=
\mathop{\rm arg\,min}\limits_{e}
\mathcal{L}_{\mathrm{rank}}
\bigl(
\mathcal{S}_u;
G+C_{c(u)}+e^{\top}h
\bigr)
+
\beta\|e\|_2^2.
\label{eq:calibration}
\end{equation}
If the new user's category is unknown, we independently fit a
user residual under every category and select the category
\(\widehat c(u)\) with the lowest regularized support loss. We
call the resulting procedure \emph{safe calibration}: the
residual is initialized from the category posterior implied by
\(\mathcal{S}_u\) rather than from zero, its norm is constrained
throughout optimization, and a budget-dependent shrinkage factor
scales the fitted residual, so smaller support sets induce more
conservative updates. The final calibrated reward is
\(
\widehat R_u(x,y)
=
G(x,y)
+
C_{\widehat c(u)}(x,y)
+
\widehat e_u^{\top}h(x,y).
\)
Adapting PALATE to a new user therefore updates neither the
global backbone nor the category residuals nor the underlying
image editor.

\section{Experiments}
\begin{table*}[t]
\centering

{\small
\setlength{\tabcolsep}{0pt}
\renewcommand{\arraystretch}{1.05}

\begin{tabular*}{\textwidth}{
@{\extracolsep{\fill}}
lcccccccccccc
@{}
}
\toprule
Method
& \shortstack{Pairwise\\Acc.}
& Top-1
& MRR
& NDCG@3
& \shortstack{Exact\\Rank}
& Hit@2
& \shortstack{Kendall\\$\tau$}
& \shortstack{Spearman\\$\rho$}
& \shortstack{Macro\\Pairwise}
& \shortstack{Worst-prof.\\Pairwise}
& \shortstack{Macro\\Top-1}
& \shortstack{Worst-prof.\\Top-1}
\\
\midrule

\multicolumn{13}{l}{\textit{Aesthetic and quality assessment}} \\

FGAesQ
& 38.31
& 21.17
& 52.24
& 73.01
& 9.17
& 49.92
& $-0.234$
& $-0.262$
& 39.60
& 35.72
& 23.08
& 17.33
\\

CLIP-IQA
& 43.72
& 25.92
& 56.92
& 75.92
& 13.42
& 59.42
& $-0.097$
& $-0.112$
& 43.94
& 42.33
& 26.21
& 24.00
\\

\midrule
\multicolumn{13}{l}{\textit{Vision--language similarity}} \\

CLIP image similarity
& 41.17
& 22.25
& 54.53
& 74.39
& 12.67
& 58.21
& $-0.148$
& $-0.165$
& 41.28
& 40.00
& 22.96
& 20.83
\\

SigLIP similarity
& 48.67
& 35.83
& 62.69
& 78.75
& 18.38
& 65.88
& 0.002
& 0.000
& 48.78
& 47.67
& 36.42
& 34.67
\\

CLIPScore
& 48.78
& 29.67
& 60.23
& 78.16
& 15.29
& 66.62
& 0.004
& 0.010
& 48.69
& 45.33
& 29.67
& 27.00
\\

\midrule
\multicolumn{13}{l}{\textit{Human-preference reward models}} \\

HPS v2
& 37.14
& 22.42
& 54.08
& 73.58
& 10.75
& 53.62
& $-0.216$
& $-0.235$
& 37.96
& 34.50
& 23.54
& 20.00
\\

PickScore
& 58.06
& 38.83
& 66.85
& 82.08
& 25.71
& 76.75
& 0.189
& 0.207
& 57.56
& 52.67
& 38.17
& 35.50
\\

\midrule
\multicolumn{13}{l}{\textit{Editing-specific reward models}} \\

\shortstack[l]{EditReward,\\generic prompt}
& 46.06
& 27.08
& 58.56
& 76.83
& 14.92
& 65.12
& $-0.051$
& $-0.044$
& 46.75
& 44.67
& 27.79
& 25.67
\\

\shortstack[l]{EditReward,\\profile-conditioned prompt}
& 48.78
& 32.33
& 61.30
& 78.31
& 18.58
& 66.42
& 0.004
& 0.009
& 48.94
& 43.50
& 32.92
& 26.50
\\

\midrule

\textbf{PALATE (ours)}
& \textbf{72.83}
& \textbf{65.08}
& \textbf{82.20}
& \textbf{90.26}
& \textbf{42.25}
& \textbf{89.96}
& \textbf{0.485}
& \textbf{0.533}
& \textbf{71.53}
& \textbf{64.67}
& \textbf{62.04}
& \textbf{52.50}
\\

\bottomrule
\end{tabular*}
}

\caption{
Comparison under the same held-out users and source images.
All metrics except Kendall's $\tau$ and Spearman's $\rho$ are
percentages. Macro scores average user profiles equally; worst-profile
scores report the lowest-performing profile. Best results are bold.
}
\label{tab:comprehensive_comparison}
\end{table*}

\subsection{Experimental Setup}
\paragraph{Baselines.}
We compare PALATE with CLIPScore \citep{hessel2021clipscore},
PickScore \citep{kirstain2023pickapic}, HPS v2
\citep{wu2023hpsv2}, SigLIP \citep{zhai2023siglip}, CLIP-IQA
\citep{wang2023clipiqa}, CLIP similarity \citep{radford2021clip},
FGAesQ \citep{yang2026fgaesq}, and Qwen2.5-VL-based
EditReward \citep{wu2026editreward,bai2025qwen25vl}. We also
evaluate global-only and wrong-category routing with the same
backbone. Few-shot calibration is compared with an unconstrained,
zero-initialized user adapter. All methods share candidate bundles,
held-out splits, support sets, and budgets
\(K\in\{0,1,5,10,20,50\}\).

\paragraph{Model Architecture.}
For each source--candidate pair, we cache a 195-dimensional
vector of color, edge, edit-difference, quality, and artifact
cues from the full image and portrait region. Its eight semantic
groups form the tokens of a two-layer Pre-LN Transformer
\citep{vaswani2017attention} with four heads and hidden size 128.
The normalized \texttt{[CLS]} embedding \(h(x,y)\) is scored by
a scalar head; the 128-dimensional user residual enters only as
\(e_u^\top h(x,y)\) in Eq.~(\ref{eq:hierarchical_reward}).

\paragraph{Datasets and Evaluation Protocol.}
Our benchmark uses 1,000 PPR10K portraits
\citep{liang2021ppr10k}, each with three expert retouches and a
portrait mask. Twenty-five volunteers rank every bundle under
five aesthetic styles, yielding 25,000 rankings and 75,000
pairwise comparisons. Users are split 17/2/6 and images
700/100/200 for joint leave-user-out and leave-image-out
training, validation, and testing; the test set contains 1,200
rankings. We report pairwise and bundle-level measures, including
NDCG@3 \citep{jarvelin2002cg}. Few-shot support is drawn only
from the 800 non-query images, and differences use unrounded
values.

\par\noindent\textbf{Training Details.} The global reward is pretrained for three epochs on
expert-retouch-versus-source comparisons. We then run five
evolution rounds: user-teacher fitting (50 epochs, 10 examples
per user), user-to-category distillation (three repetitions of
two epochs), category-to-global consolidation (two epochs), and
global broadcast (one epoch). We select models by zero-shot
validation accuracy and run all experiments on four NVIDIA A800 GPUs.

\begin{figure}[t]
\centering
\includegraphics[width=0.86\columnwidth]{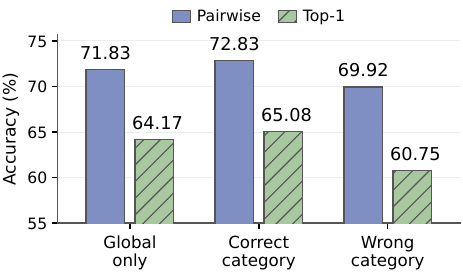}
\caption{Deployment modes under the user--image double-held-out protocol.
Correct routing improves global-only scoring; wrong routing degrades it.}
\label{fig:deployment_modes}
\end{figure}

\begin{figure}[t]
\noindent
\includegraphics[width=0.82\columnwidth]{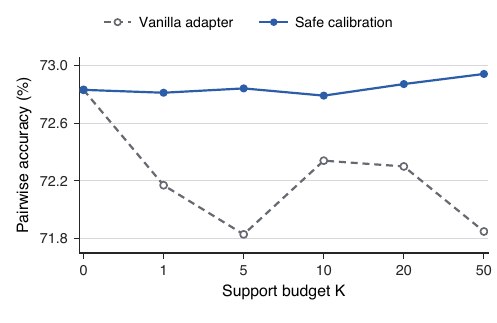}
\caption{
Few-shot user calibration under different support budgets.
Safe calibration is stable; the vanilla adapter degrades.
}
\label{fig:fewshot}
\end{figure}

\subsection{Comparison with Existing Reward Models}
PALATE leads every scorer under the user--image double-held-out
protocol (Table~\ref{tab:comprehensive_comparison}), consistent
with the selections in Figure~\ref{fig:qualitative_comparison}.
It reaches 72.83\% pairwise and 65.08\% Top-1 accuracy; the
strongest baseline, PickScore, reaches 58.06\% pairwise
accuracy. Profile-conditioned EditReward improves over its
generic prompt (48.78\% versus 46.06\%) but remains 24.05
points behind PALATE. Textual profiles therefore help, yet do
not replace task-aligned preference evidence accumulated across
users.

\subsection{Effect of Hierarchical Preference Modeling}
The global reward obtains 71.83\% pairwise and 64.17\% Top-1
accuracy (Figure~\ref{fig:deployment_modes}). Correct routing
raises them to 72.83\% and 65.08\%, whereas wrong routing
reduces them to 69.92\% and 60.75\%. The 2.92-point routing
gap has an image-cluster bootstrap \citep{efron1979bootstrap}
95\% interval of $[1.50,4.39]$ and an exact McNemar
\citep{mcnemar1947sampling} \(p=8.24\times10^{-7}\).
Category rewards thus encode reusable group preference rather
than arbitrary extra capacity.

\subsection{Safe Few-Shot User Calibration}
The vanilla adapter falls from 72.83\% at \(K=0\) to 71.85\%
at \(K=50\), while safe calibration remains stable and reaches
72.94\% (Figure~\ref{fig:fewshot}). Its \(K=50\) gain over
zero-shot is 0.111 points, with a user-cluster bootstrap 95\%
interval of \([0.019,0.204]\). Category-posterior
initialization, norm constraints, and budget-dependent
shrinkage therefore turn additional feedback into a small but
reliable gain rather than overfitting.

\begin{table}[t]
\centering
\small
\setlength{\tabcolsep}{3pt}
\begin{tabular}{@{}lrr@{}}
\toprule
\multicolumn{3}{l}{\textit{Persistent storage}} \\
Component & Parameters & Storage \\
\midrule
Global reward & 309,511 & 1.23 MiB \\
Five category rewards & $5\times309{,}511$ & 6.08 MiB \\
One user adapter & 128 & 512 B \\
\midrule
\multicolumn{3}{l}{\textit{Computation time}} \\
Operation & Time & Setting \\
\midrule
Global scoring & $2.18\pm0.64$ ms & 0.189 ms/cand.\ ($B=8$) \\
Global + category & $3.31\pm0.60$ ms & 0.488 ms/cand.\ ($B=8$) \\
Direct adaptation & 1.99 s/user & 1.44--3.95 s range \\
\bottomrule
\end{tabular}
\caption{Storage and deployment cost measured on one NVIDIA
A800-SXM4-80GB GPU with cached 195-dimensional input features.
The shared PALATE state, consisting of one global and five category
rewards, occupies 7.31 MiB. The adaptation timing is measured for the
vanilla user adapter at $K=10$ with 50 optimization steps and is
reported as its mean and observed range over six test users.}
\label{tab:efficiency}
\end{table}

\begin{table}[t]
\centering
\small
\renewcommand{\arraystretch}{1.02}
\begin{tabular}{lc}
\toprule
Variant & Pairwise acc. (\%) \\
\midrule
Full PALATE & \textbf{72.83} \\
\midrule
\multicolumn{2}{l}{\textit{Cyclic knowledge flow}} \\
w/o user-to-category distillation & 71.50 \\
w/o category-to-global distillation & 72.72 \\
w/o global broadcast & 72.03 \\
w/o double teacher & 72.33 \\
One inner step & 72.17 \\
Reset category rewards & 72.61 \\
Shuffled categories & 70.14 \\
\midrule
\multicolumn{2}{l}{\textit{Input representations}} \\
w/o source tokens & 70.89 \\
w/o edit tokens & 70.17 \\
w/o portrait tokens & 72.03 \\
\midrule
\multicolumn{2}{l}{\textit{Anti-collapse regularization}} \\
w/o category anti-collapse & 72.06 \\
w/o both anti-collapse terms & 71.25 \\
\bottomrule
\end{tabular}
\caption{Ablation results under the same user--image double-held-out
protocol. All variants are selected using the validation set.
The three blocks evaluate cyclic knowledge flow, input
representations, and anti-collapse regularization, respectively.
Pairwise accuracy is reported in percent.}
\label{tab:ablation}
\end{table}

\subsection{Deployment Efficiency}
PALATE stores one user in 128 FP32 values (512 B), complementing
parameter-efficient adaptation \citep{houlsby2019adapter,hu2022lora}
with substantially smaller persistent state
(Table~\ref{tab:efficiency}). The shared rewards occupy
7.31 MiB. Global-plus-category scoring costs
\(3.31\pm0.60\) ms per candidate at batch size one and
0.488 ms at batch size eight, keeping a three-candidate bundle
below two milliseconds at the latter setting. With 10,000 users,
the shared rewards and all user states require 12.19 MiB; even
one million adapters remain below 500 MiB. Personalization
therefore scales with users through byte-scale states rather
than duplicated reward models.

\subsection{Ablation Study}
Category semantics and source/edit cues are the dominant factors
(Table~\ref{tab:ablation}). Removing user-to-category
distillation loses 1.33 points (95\% image-cluster interval
\([0.50,2.22]\)); shuffling categories loses 2.69 points,
more than resetting category rewards (0.22 points). Hence the
group semantics matter more than the initial residual values.
Removing source or edit tokens reduces accuracy to 70.89\% and
70.17\%, respectively, while removing both anti-collapse terms
yields 71.25\%. Category-to-global distillation, global broadcast, and the double teacher each contribute stably, with changes of at most 0.81 points. These results locate the primary gains in meaningful preference groups, edit-aware evidence, and stable representation learning, with cyclic refinements providing effective complementary improvements. Portrait-mask tokens also provide consistent benefits: removing them decreases accuracy to 72.03\%,indicating that global cues explain most rankings while localized
face and skin evidence remains additionally useful. Category
anti-collapse alone contributes 0.77 points, and using both
anti-collapse terms together yields a total improvement of 1.58
points, demonstrating their complementary roles. Among cyclic
components, user-to-category transfer is the most influential,
effectively distilling individual supervision into reusable category
knowledge.

\section{Conclusion}
PALATE personalizes portrait-retouch selection through a
global--category--user reward hierarchy that cyclically shares
knowledge while retaining a 512-byte state per user. Without
modifying the editor, anti-collapse regularization preserves
distinct preference geometry. On the user--image
double-held-out PPR10K benchmark, PALATE reaches 72.83\%
pairwise accuracy and outperforms all baselines. Routing and
ablations attribute the gain to meaningful preference groups,
while constrained calibration safely incorporates few-shot
feedback. Together with millisecond-level scoring, this
combination of shared evolution and byte-scale user state offers
a practical design for subjective visual tasks in which separate
per-user models are infeasible. Future work will infer categories
from feedback and couple the reward with candidate generation.

\begingroup
\small
\bibliography{references}

@article{zhang2026aiedit,
  author  = {Zhang, Peixuan and Weng, Shuchen and Zhu, Chengxuan and Tang, Binghao and Jia, Zijian and Li, Si and Shi, Boxin},
  title   = {Affective Image Editing: Shaping Emotional Factors via Text Descriptions},
  journal = {International Journal of Computer Vision},
  volume  = {134},
  number  = {1},
  pages   = {16},
  year    = {2026},
  doi     = {10.1007/s11263-025-02678-y}
}

@misc{zhu2025pif,
  author        = {Zhu, Chengxuan and Weng, Shuchen and Fang, Jiacong and Zhang, Peixuan and Li, Si and Xu, Chao and Shi, Boxin},
  title         = {Personalized Image Filter: Mastering Your Photographic Style},
  year          = {2025},
  eprint        = {2510.16791},
  archivePrefix = {arXiv}
}

@inproceedings{moran2020deeplpf,
  author    = {Moran, Sean and Marza, Pierre and McDonagh, Steven and Parisot, Sarah and Slabaugh, Gregory},
  title     = {{DeepLPF}: Deep Local Parametric Filters for Image Enhancement},
  booktitle = {Proceedings of the IEEE/CVF Conference on Computer Vision and Pattern Recognition (CVPR)},
  pages     = {12826--12835},
  year      = {2020}
}

@inproceedings{liang2021ppr10k,
  author    = {Liang, Jie and Zeng, Hui and Cui, Miaomiao and Xie, Xuansong and Zhang, Lei},
  title     = {{PPR10K}: A Large-Scale Portrait Photo Retouching Dataset with Human-Region Mask and Group-Level Consistency},
  booktitle = {Proceedings of the IEEE/CVF Conference on Computer Vision and Pattern Recognition (CVPR)},
  pages     = {653--661},
  year      = {2021}
}

@inproceedings{ouyang2023rsfnet,
  author    = {Ouyang, Wenqi and Dong, Yi and Kang, Xiaoyang and Ren, Peiran and Xu, Xin and Xie, Xuansong},
  title     = {{RSFNet}: A White-Box Image Retouching Approach Using Region-Specific Color Filters},
  booktitle = {Proceedings of the IEEE/CVF International Conference on Computer Vision (ICCV)},
  pages     = {12160--12169},
  year      = {2023}
}

@inproceedings{xie2023bpfre,
  author    = {Xie, Lianxin and Xue, Wen and Xu, Zhen and Wu, Si and Yu, Zhiwen and Wong, Hau San},
  title     = {Blemish-Aware and Progressive Face Retouching with Limited Paired Data},
  booktitle = {Proceedings of the IEEE/CVF Conference on Computer Vision and Pattern Recognition (CVPR)},
  pages     = {5599--5608},
  year      = {2023}
}

@inproceedings{bychkovsky2011fivek,
  author    = {Bychkovsky, Vladimir and Paris, Sylvain and Chan, Eric and Durand, Fr{\'e}do},
  title     = {Learning Photographic Global Tonal Adjustment with a Database of Input/Output Image Pairs},
  booktitle = {Proceedings of the IEEE Conference on Computer Vision and Pattern Recognition (CVPR)},
  year      = {2011}
}

@inproceedings{xu2023imagereward,
  author    = {Xu, Jiazheng and Liu, Xiao and Wu, Yuchen and Tong, Yuxuan and Li, Qinkai and Ding, Ming and Tang, Jie and Dong, Yuxiao},
  title     = {{ImageReward}: Learning and Evaluating Human Preferences for Text-to-Image Generation},
  booktitle = {Advances in Neural Information Processing Systems},
  volume    = {36},
  year      = {2023}
}

@inproceedings{wu2023hps,
  author    = {Wu, Xiaoshi and Sun, Keqiang and Zhu, Feng and Zhao, Rui and Li, Hongsheng},
  title     = {Human Preference Score: Better Aligning Text-to-Image Models with Human Preference},
  booktitle = {Proceedings of the IEEE/CVF International Conference on Computer Vision (ICCV)},
  pages     = {2096--2105},
  year      = {2023}
}

@misc{wu2023hpsv2,
  author        = {Wu, Xiaoshi and Hao, Yiming and Sun, Keqiang and Chen, Yixiong and Zhu, Feng and Zhao, Rui and Li, Hongsheng},
  title         = {Human Preference Score v2: A Solid Benchmark for Evaluating Human Preferences of Text-to-Image Synthesis},
  year          = {2023},
  eprint        = {2306.09341},
  archivePrefix = {arXiv}
}

@inproceedings{zhang2024mps,
  author    = {Zhang, Sixian and Wang, Bohan and Wu, Junqiang and Li, Yan and Gao, Tingting and Zhang, Di and Wang, Zhongyuan},
  title     = {Learning Multi-Dimensional Human Preference for Text-to-Image Generation},
  booktitle = {Proceedings of the IEEE/CVF Conference on Computer Vision and Pattern Recognition (CVPR)},
  pages     = {8018--8027},
  year      = {2024}
}

@inproceedings{zhang2024hive,
  author    = {Zhang, Shu and Yang, Xinyi and Feng, Yihao and Qin, Can and Chen, Chia-Chih and Yu, Ning and Chen, Zeyuan and Wang, Huan and Savarese, Silvio and Ermon, Stefano and Xiong, Caiming and Xu, Ran},
  title     = {{HIVE}: Harnessing Human Feedback for Instructional Visual Editing},
  booktitle = {Proceedings of the IEEE/CVF Conference on Computer Vision and Pattern Recognition (CVPR)},
  pages     = {9026--9036},
  year      = {2024}
}

@inproceedings{kirstain2023pickapic,
  author    = {Kirstain, Yuval and Polyak, Adam and Singer, Uriel and Matiana, Shahbuland and Penna, Joe and Levy, Omer},
  title     = {Pick-a-Pic: An Open Dataset of User Preferences for Text-to-Image Generation},
  booktitle = {Advances in Neural Information Processing Systems},
  volume    = {36},
  year      = {2023}
}

@inproceedings{wu2026editreward,
  author    = {Wu, Keming and Jiang, Sicong and Ku, Max and Nie, Ping and Liu, Minghao and Chen, Wenhu},
  title     = {{EditReward}: A Human-Aligned Reward Model for Instruction-Guided Image Editing},
  booktitle = {Proceedings of the 14th International Conference on Learning Representations (ICLR)},
  year      = {2026}
}

@inproceedings{ren2017personalized,
  author    = {Ren, Jian and Shen, Xiaohui and Lin, Zhe and M{\v{e}}ch, Radom{\'i}r and Foran, David J.},
  title     = {Personalized Image Aesthetics},
  booktitle = {Proceedings of the IEEE International Conference on Computer Vision (ICCV)},
  pages     = {638--647},
  year      = {2017}
}

@inproceedings{yang2022para,
  author    = {Yang, Yuzhe and Xu, Liwu and Li, Leida and Qie, Nan and Li, Yaqian and Zhang, Peng and Guo, Yandong},
  title     = {Personalized Image Aesthetics Assessment with Rich Attributes},
  booktitle = {Proceedings of the IEEE/CVF Conference on Computer Vision and Pattern Recognition (CVPR)},
  pages     = {19861--19869},
  year      = {2022}
}

@inproceedings{dang2025personalized,
  author    = {Dang, Meihua and Singh, Anikait and Zhou, Linqi and Ermon, Stefano and Song, Jiaming},
  title     = {Personalized Preference Fine-tuning of Diffusion Models},
  booktitle = {Proceedings of the IEEE/CVF Conference on Computer Vision and Pattern Recognition (CVPR)},
  year      = {2025}
}

@inproceedings{yang2026fgaesq,
  author    = {Yang, Zhichao and Wang, Jianjie and Zhang, Zhixianhe and Xie, Pangu and Sheng, Xiangfei and Chen, Pengfei and Li, Leida},
  title     = {Fine-Grained Image Aesthetic Assessment: Learning Discriminative Scores from Relative Ranks},
  booktitle = {Proceedings of the IEEE/CVF Conference on Computer Vision and Pattern Recognition (CVPR)},
  year      = {2026}
}

@inproceedings{poddar2024vpl,
  author    = {Poddar, Sriyash and Wan, Yanming and Ivison, Hamish and Gupta, Abhishek and Jaques, Natasha},
  title     = {Personalizing Reinforcement Learning from Human Feedback with Variational Preference Learning},
  booktitle = {Advances in Neural Information Processing Systems},
  volume    = {37},
  year      = {2024}
}

@misc{bose2025lore,
  author        = {Bose, Avinandan and Xiong, Zhihan and Chi, Yuejie and Du, Simon Shaolei and Xiao, Lin and Fazel, Maryam},
  title         = {{LoRe}: Personalizing {LLMs} via Low-Rank Reward Modeling},
  year          = {2025},
  eprint        = {2504.14439},
  archivePrefix = {arXiv}
}

@misc{hinton2015distilling,
  author        = {Hinton, Geoffrey and Vinyals, Oriol and Dean, Jeff},
  title         = {Distilling the Knowledge in a Neural Network},
  year          = {2015},
  eprint        = {1503.02531},
  archivePrefix = {arXiv}
}

@inproceedings{zhang2018dml,
  author    = {Zhang, Ying and Xiang, Tao and Hospedales, Timothy M. and Lu, Huchuan},
  title     = {Deep Mutual Learning},
  booktitle = {Proceedings of the IEEE Conference on Computer Vision and Pattern Recognition (CVPR)},
  pages     = {4320--4328},
  year      = {2018}
}

@inproceedings{furlanello2018born,
  author    = {Furlanello, Tommaso and Lipton, Zachary and Tschannen, Michael and Itti, Laurent and Anandkumar, Anima},
  title     = {Born Again Neural Networks},
  booktitle = {Proceedings of the 35th International Conference on Machine Learning},
  volume    = {80},
  pages     = {1607--1616},
  publisher = {PMLR},
  year      = {2018}
}

@inproceedings{lin2020feddf,
  author    = {Lin, Tao and Kong, Lingjing and Stich, Sebastian U. and Jaggi, Martin},
  title     = {Ensemble Distillation for Robust Model Fusion in Federated Learning},
  booktitle = {Advances in Neural Information Processing Systems},
  volume    = {33},
  year      = {2020}
}

@inproceedings{son2021densely,
  author    = {Son, Wonchul and Na, Jaemin and Choi, Junyong and Hwang, Wonjun},
  title     = {Densely Guided Knowledge Distillation Using Multiple Teacher Assistants},
  booktitle = {Proceedings of the IEEE/CVF International Conference on Computer Vision (ICCV)},
  pages     = {9395--9404},
  year      = {2021}
}

@inproceedings{choi2023orc,
  author    = {Choi, Junyong and Cho, Hyeon and Cheung, Seokhwa and Hwang, Wonjun},
  title     = {{ORC}: Network Group-Based Knowledge Distillation Using Online Role Change},
  booktitle = {Proceedings of the IEEE/CVF International Conference on Computer Vision (ICCV)},
  pages     = {17381--17390},
  year      = {2023}
}

@inproceedings{taha2021knowledge,
  author    = {Taha, Ahmed and Shrivastava, Abhinav and Davis, Larry S.},
  title     = {Knowledge Evolution in Neural Networks},
  booktitle = {Proceedings of the IEEE/CVF Conference on Computer Vision and Pattern Recognition (CVPR)},
  pages     = {12843--12852},
  year      = {2021}
}

@inproceedings{wen2024cilmtd,
  author    = {Wen, Haitao and Pan, Lili and Dai, Yu and Qiu, Heqian and Wang, Lanxiao and Wu, Qingbo and Li, Hongliang},
  title     = {Class Incremental Learning with Multi-Teacher Distillation},
  booktitle = {Proceedings of the IEEE/CVF Conference on Computer Vision and Pattern Recognition (CVPR)},
  pages     = {28443--28452},
  year      = {2024}
}

@inproceedings{hu2022lora,
  author    = {Hu, Edward J. and Shen, Yelong and Wallis, Phillip and Allen-Zhu, Zeyuan and Li, Yuanzhi and Wang, Shean and Wang, Lu and Chen, Weizhu},
  title     = {{LoRA}: Low-Rank Adaptation of Large Language Models},
  booktitle = {Proceedings of the 10th International Conference on Learning Representations (ICLR)},
  year      = {2022}
}

@inproceedings{vaswani2017attention,
  author    = {Vaswani, Ashish and Shazeer, Noam and Parmar, Niki and Uszkoreit, Jakob and Jones, Llion and Gomez, Aidan N. and Kaiser, {\L}ukasz and Polosukhin, Illia},
  title     = {Attention Is All You Need},
  booktitle = {Advances in Neural Information Processing Systems},
  volume    = {30},
  year      = {2017}
}

@inproceedings{radford2021clip,
  author    = {Radford, Alec and Kim, Jong Wook and Hallacy, Chris and Ramesh, Aditya and Goh, Gabriel and Agarwal, Sandhini and Sastry, Girish and Askell, Amanda and Mishkin, Pamela and Clark, Jack and Krueger, Gretchen and Sutskever, Ilya},
  title     = {Learning Transferable Visual Models from Natural Language Supervision},
  booktitle = {Proceedings of the 38th International Conference on Machine Learning},
  volume    = {139},
  pages     = {8748--8763},
  publisher = {PMLR},
  year      = {2021}
}

@inproceedings{hessel2021clipscore,
  author    = {Hessel, Jack and Holtzman, Ari and Forbes, Maxwell and Le Bras, Ronan and Choi, Yejin},
  title     = {{CLIPScore}: A Reference-Free Evaluation Metric for Image Captioning},
  booktitle = {Proceedings of the 2021 Conference on Empirical Methods in Natural Language Processing (EMNLP)},
  pages     = {7514--7528},
  year      = {2021}
}

@inproceedings{zhai2023siglip,
  author    = {Zhai, Xiaohua and Mustafa, Basil and Kolesnikov, Alexander and Beyer, Lucas},
  title     = {Sigmoid Loss for Language Image Pre-Training},
  booktitle = {Proceedings of the IEEE/CVF International Conference on Computer Vision (ICCV)},
  pages     = {11975--11986},
  year      = {2023}
}

@article{wang2023clipiqa,
  author  = {Wang, Jianyi and Chan, Kelvin C. K. and Loy, Chen Change},
  title   = {Exploring {CLIP} for Assessing the Look and Feel of Images},
  journal = {Proceedings of the AAAI Conference on Artificial Intelligence},
  volume  = {37},
  number  = {2},
  pages   = {2555--2563},
  year    = {2023}
}

@misc{balestriero2025lejepa,
  author        = {Balestriero, Randall and LeCun, Yann},
  title         = {{LeJEPA}: Provable and Scalable Self-Supervised Learning Without the Heuristics},
  year          = {2025},
  eprint        = {2511.08544},
  archivePrefix = {arXiv}
}

@article{bradley1952rank,
  author  = {Bradley, Ralph Allan and Terry, Milton E.},
  title   = {Rank Analysis of Incomplete Block Designs: {I}. The Method of Paired Comparisons},
  journal = {Biometrika},
  volume  = {39},
  number  = {3/4},
  pages   = {324--345},
  year    = {1952}
}

@article{epps1983test,
  author  = {Epps, T. W. and Pulley, L. B.},
  title   = {A Test for Normality Based on the Empirical Characteristic Function},
  journal = {Biometrika},
  volume  = {70},
  number  = {3},
  pages   = {723--726},
  year    = {1983}
}

@article{gharbi2017hdrnet,
  author    = {Gharbi, Micha{\"e}l and Chen, Jiawen and Barron, Jonathan T. and Hasinoff, Samuel W. and Durand, Fr{\'e}do},
  title     = {Deep Bilateral Learning for Real-Time Image Enhancement},
  journal   = {ACM Transactions on Graphics},
  volume    = {36},
  number    = {4},
  year      = {2017}
}

@article{hu2018exposure,
  author    = {Hu, Yuanming and He, Hao and Xu, Chenxi and Wang, Baoyuan and Lin, Stephen},
  title     = {Exposure: A White-Box Photo Post-Processing Framework},
  journal   = {ACM Transactions on Graphics},
  volume    = {37},
  number    = {2},
  year      = {2018}
}

@article{zeng2022lut,
  author    = {Zeng, Hui and Cai, Jianrui and Li, Lida and Cao, Zisheng and Zhang, Lei},
  title     = {Learning Image-Adaptive {3D} Lookup Tables for High Performance Photo Enhancement in Real-Time},
  journal   = {IEEE Transactions on Pattern Analysis and Machine Intelligence},
  volume    = {44},
  number    = {4},
  pages     = {2058--2073},
  year      = {2022}
}

@inproceedings{he2020csrnet,
  author    = {He, Jingwen and Liu, Yihao and Qiao, Yu and Dong, Chao},
  title     = {Conditional Sequential Modulation for Efficient Global Image Retouching},
  booktitle = {Proceedings of the European Conference on Computer Vision (ECCV)},
  year      = {2020}
}

@inproceedings{brooks2023instructpix2pix,
  author    = {Brooks, Tim and Holynski, Aleksander and Efros, Alexei A.},
  title     = {{InstructPix2Pix}: Learning to Follow Image Editing Instructions},
  booktitle = {Proceedings of the IEEE/CVF Conference on Computer Vision and Pattern Recognition (CVPR)},
  pages     = {18392--18402},
  year      = {2023}
}

@inproceedings{zhang2023magicbrush,
  author    = {Zhang, Kai and Mo, Lingbo and Chen, Wenhu and Sun, Huan and Su, Yu},
  title     = {{MagicBrush}: A Manually Annotated Dataset for Instruction-Guided Image Editing},
  booktitle = {Advances in Neural Information Processing Systems, Datasets and Benchmarks Track},
  volume    = {36},
  year      = {2023}
}

@article{zhu2022blgpiaa,
  author    = {Zhu, Hancheng and Li, Leida and Wu, Jinjian and Zhao, Sicheng and Ding, Guiguang and Shi, Guangming},
  title     = {Personalized Image Aesthetics Assessment via Meta-Learning with Bilevel Gradient Optimization},
  journal   = {IEEE Transactions on Cybernetics},
  volume    = {52},
  number    = {3},
  pages     = {1798--1811},
  year      = {2022}
}

@inproceedings{houlsby2019adapter,
  author    = {Houlsby, Neil and Giurgiu, Andrei and Jastrzebski, Stanislaw and Morrone, Bruna and De Laroussilhe, Quentin and Gesmundo, Andrea and Attariyan, Mona and Gelly, Sylvain},
  title     = {Parameter-Efficient Transfer Learning for {NLP}},
  booktitle = {Proceedings of the 36th International Conference on Machine Learning},
  volume    = {97},
  pages     = {2790--2799},
  publisher = {PMLR},
  year      = {2019}
}

@inproceedings{romero2015fitnets,
  author    = {Romero, Adriana and Ballas, Nicolas and Kahou, Samira Ebrahimi and Chassang, Antoine and Gatta, Carlo and Bengio, Yoshua},
  title     = {{FitNets}: Hints for Thin Deep Nets},
  booktitle = {Proceedings of the 3rd International Conference on Learning Representations (ICLR)},
  year      = {2015}
}

@article{papyan2020collapse,
  author    = {Papyan, Vardan and Han, X. Y. and Donoho, David L.},
  title     = {Prevalence of Neural Collapse during the Terminal Phase of Deep Learning Training},
  journal   = {Proceedings of the National Academy of Sciences},
  volume    = {117},
  number    = {40},
  pages     = {24652--24663},
  year      = {2020}
}

@misc{bai2025qwen25vl,
  author        = {Bai, Shuai and Chen, Keqin and Liu, Xuejing and Wang, Jialin and Ge, Wenbin and Song, Sibo and Dang, Kai and Wang, Peng and Wang, Shijie and Tang, Jun and Zhong, Humen and Zhu, Yuanzhi and Yang, Mingkun and Li, Zhaohai and Wan, Jianqiang and Wang, Pengfei and Ding, Wei and Fu, Zheren and Xu, Yiheng and Ye, Jiabo and Zhang, Xi and Xie, Tianbao and Cheng, Zesen and Zhang, Hang and Yang, Zhibo and Xu, Haiyang and Lin, Junyang},
  title         = {{Qwen2.5-VL} Technical Report},
  year          = {2025},
  eprint        = {2502.13923},
  archivePrefix = {arXiv}
}

@article{jarvelin2002cg,
  author  = {J{\"a}rvelin, Kalervo and Kek{\"a}l{\"a}inen, Jaana},
  title   = {Cumulated Gain-Based Evaluation of {IR} Techniques},
  journal = {ACM Transactions on Information Systems},
  volume  = {20},
  number  = {4},
  pages   = {422--446},
  year    = {2002},
  doi     = {10.1145/582415.582418}
}

@article{efron1979bootstrap,
  author  = {Efron, Bradley},
  title   = {Bootstrap Methods: Another Look at the Jackknife},
  journal = {The Annals of Statistics},
  volume  = {7},
  number  = {1},
  pages   = {1--26},
  year    = {1979},
  doi     = {10.1214/aos/1176344552}
}

@article{mcnemar1947sampling,
  author  = {McNemar, Quinn},
  title   = {Note on the Sampling Error of the Difference Between Correlated Proportions or Percentages},
  journal = {Psychometrika},
  volume  = {12},
  number  = {2},
  pages   = {153--157},
  year    = {1947},
  doi     = {10.1007/BF02295996}
}

@inproceedings{weng2020cirbridge,
  author    = {Weng, Shuchen and Li, Wenbo and Li, Dawei and Jin, Hongxia and Shi, Boxin},
  title     = {Conditional Image Repainting via Semantic Bridge and Piecewise Value Function},
  booktitle = {Proceedings of the European Conference on Computer Vision (ECCV)},
  pages     = {467--482},
  year      = {2020}
}

@inproceedings{sun2022unicorn,
  author    = {Sun, Jimeng and Weng, Shuchen and Chang, Zheng and Li, Si and Shi, Boxin},
  title     = {{UniCoRN}: A Unified Conditional Image Repainting Network},
  booktitle = {Proceedings of the IEEE/CVF Conference on Computer Vision and Pattern Recognition (CVPR)},
  pages     = {11369--11378},
  year      = {2022}
}

@inproceedings{tang2023luminaire,
  author    = {Tang, Jiajun and Zhong, Haofeng and Weng, Shuchen and Shi, Boxin},
  title     = {{LuminAIRe}: Illumination-Aware Conditional Image Repainting for Lighting-Realistic Generation},
  booktitle = {Advances in Neural Information Processing Systems},
  volume    = {36},
  pages     = {64468--64481},
  year      = {2023}
}

@article{weng2024cir,
  author  = {Weng, Shuchen and Shi, Boxin},
  title   = {Conditional Image Repainting},
  journal = {IEEE Transactions on Pattern Analysis and Machine Intelligence},
  volume  = {46},
  number  = {4},
  pages   = {2285--2298},
  year    = {2024}
}

@article{weng2025opencir,
  author  = {Weng, Shuchen and Gong, Xiaocheng and Zheng, Haojie and Wang, Xinlong and Li, Si and Shi, Boxin},
  title   = {{OpenCIR}: Conditional Image Repainting with Open Condition Mixture},
  journal = {IEEE Transactions on Pattern Analysis and Machine Intelligence},
  volume  = {47},
  number  = {11},
  pages   = {10406--10419},
  year    = {2025}
}

@inproceedings{weng2023aif,
  author    = {Weng, Shuchen and Zhang, Peixuan and Chang, Zheng and Wang, Xinlong and Li, Si and Shi, Boxin},
  title     = {Affective Image Filter: Reflecting Emotions from Text to Images},
  booktitle = {Proceedings of the IEEE/CVF International Conference on Computer Vision (ICCV)},
  pages     = {10810--10819},
  year      = {2023}
}

@article{zhang2026aifd,
  author  = {Zhang, Peixuan and Weng, Shuchen and Tang, Jiajun and Li, Si and Shi, Boxin},
  title   = {Toward Deeper Emotional Reflection: Crafting Affective Image Filters with Generative Priors},
  journal = {IEEE Transactions on Pattern Analysis and Machine Intelligence},
  volume  = {48},
  number  = {4},
  pages   = {4303--4317},
  year    = {2026}
}
\endgroup
\end{document}